\documentclass{article}
\usepackage{spconf,amsmath,graphicx}
\usepackage{multirow} 
\usepackage{amssymb}

\usepackage{setspace}

\title{ROBUST SINGLE-PARTICLE CRYO-EM IMAGE DENOISING and RESTORATION}
\name{Jing Zhang\textsuperscript{1,2,*}, Tengfei Zhao\textsuperscript{1,*}, ShiYu Hu\textsuperscript{1}, Xin Zhao\textsuperscript{2,\dag}}
\address{1. Institute of Automation, Chinese Academy of Sciences, Beijing, China\\
2. School of Computer and Communication Engineering\\ University of Science and Technology Beijing, Beijing, China}

\begin{document}
%
\maketitle

\renewcommand{\thefootnote}{*}
\footnotetext{Equal Contribution, \dag Corresponding Author}

%
\begin{abstract}
Cryo-electron microscopy~(cryo-EM) has achieved \textit{near-atomic level} resolution of biomolecules by reconstructing 2D micrographs. However, the resolution and accuracy of the reconstructed particles are significantly reduced due to the extremely low signal-to-noise ratio (SNR) and complex noise structure of cryo-EM images.~In this paper, we introduce a diffusion model with post-processing framework to effectively denoise and restore single particle cryo-EM images. Our method outperforms the state-of-the-art (SOTA) denoising methods by effectively removing structural noise that has not been addressed before. Additionally, more accurate and high-resolution three-dimensional reconstruction structures can be obtained from denoised cryo-EM images.
\end{abstract} 
\vspace{-0.3cm} 
\section{Introduction}
\label{sec:intro}
\vspace{-0.2cm} 
Cryo-electron microscopy (cryo-EM), as a crucial research tool in structural biology, enables the determination of molecular structures with high resolution, which relies on the accurate interpretation of particle structures \cite{wu2021machine}. However, due to the limitations of low electron dose, 2D cryo-EM images always suffer from extremely low signal-to-noise ratio (SNR) (see Fig.~1(a)), and the presence of complex noise can result in inaccurate protein structure analysis. Currently, the advancement of computer vision has infused new vitality into various cryo-EM tasks, including image denoising and restoration \cite{bhamre2016denoising} \cite{marshall2020image}, image clustering\cite{ji2018itervm}, particle picking \cite{bepler2019positive}, and more. 

The basic task of cryo-EM image processing is to minimize the interference of randomly distributed background noise, and restores the frozen state of particles in glassy ice. One representative tool for this purpose is Topaz-Denoise \cite{bepler2020topaz}, which is publicly available denoising model specifically designed for cryoEM images. It leverages the Noise2Noise framework \cite{lehtinen2018noise2noise} to directly learn the denoising from the original data,  thereby improving the SNR and accelerating particle picking. The following 2-D clustering is also designed to generate high-quality particle class averages through operations such as grouping and alignment. It has a certain guiding effect on the evaluation of data quality and subsequent 3D reconstruction \cite{fan2021cryo}. However, cryo-EM images contain not only the aforementioned randomly distributed background noise but also structural noise caused by the presence of carbon and ice in the background, as well as variations in molecular conformation \cite{baxter2009determination}.~Traditional image processing methods like Block-matching and 3-D filtering (BM3D) \cite{dabov2007image} and K-SVD \cite{aharon2006k} are very difficult to restore clear particles in high background noise, and it is more challenging to achieve structural noise reduction.~As demonstrated in Figs.~1(b) and~1(c), where the reconstructed 3D structure of the human HCN1 hyperpolarization-activated cyclic nucleotide-gated ion channel still exhibits indistinguishable structural noise when compared with the original biological structure. 
\begin{figure}[t]
	\centering
    \includegraphics[width=\linewidth]{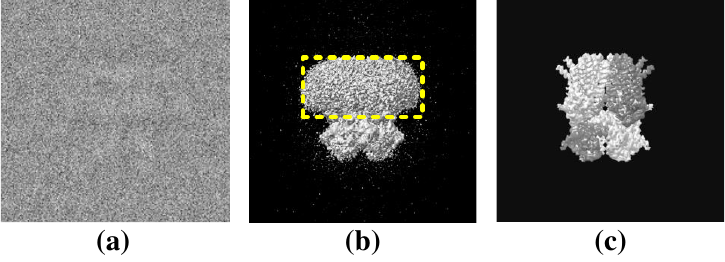} 
	\caption{Noises cryo-EM images of the HCN1 channel\cite{lee2017structures}. \textbf{(a)} Real cryo-EM image (EMPIAR-10081).~\textbf{(b)} 3D cryo-EM density map (EMD-8511, yellow box represents structural noise). \textbf{(c)} Biological structure (PDB: 5u6o). }
	\label{fig:oct}
 \vspace{-0.4cm} 
\end{figure}

In recent years, the continuous improvement of computing power and advancements in learning algorithms have made it possible to analyze complex noise.~The Noise-Transfer2Clean (NT2C) \cite{li2022noise} claims to be the first denoising method to address complex noise for the cryo-EM images. The core idea is using Generative Adversarial Networks (GAN) to learn the noise model of cryo-EM images, and further design a contrast guided noise and signal re-weighted algorithm to solve complex noise. In order to solve the problem that traditional image generation models may be polluted by a small portion of unknown outliers, Hanlin Gu et al. \cite{gu2020generative} utilized $\beta$-GAN achieves robust estimation of certain distribution parameters. It can be noticed that the strong capabilities of generative networks in the denoising and restoration of cryo-EM images due to their unique technical characteristics.

At present, the diffusion model \cite{ho2020denoising}, as a new branch of the generative networks, has shown brilliant results in natural image processing. Moreover, diffusion model also significantly surpasses GANs in terms of interpretability and stability. Furthermore, the promising development of DeepTracer \cite{guan2022deeptracer} is encouraging, which currently achieves 3D electron density map denoising by utilizing the patterns of noise and the biological structure. This approach effectively eliminates noise signals, including both background noise and structural noise. Inspired by the achievements of DeepTracer, we aim to explore future research directions in single-particle denoising and restoration.~The goal is to directly address complex noise in cryo-EM images and accurately restore the true structure of the particles. By directly improving the image quality, many tedious follow-up works can be circumvented, leading to superior reconstruction outcomes.

In this study, we present a framework that combines the diffusion model with post-processing techniques to address the challenging task of single-particle cryo-EM images denoising and restoring. Our contributions are listed as follows:
\begin{itemize}

\item Our method offers a direct approach for denoising and restoring single-particle cryo-EM images in situations of extremely low signal-to-noise ratio (SNR).

\item Complex noise in cryo-EM images can be effectively removed by our method, especially structural noise. It is worth noting that no existing method addresses structural noise in real cryo-EM images.

\item Our method achieves the best denoising effect on both simulated and real datasets. By leveraging the denoised cryo-EM images, we can attain more accurate and higher-resolution 3D structures.

\end{itemize}

\vspace{-0.4cm} 
\section{METHOD}
\label{sec:majhead}
\vspace{-0.1cm} 

\subsection{Architecture Design}
\label{ssec:subhead}
\begin{figure}[t]
	\centering
    \includegraphics[width=0.95\linewidth]{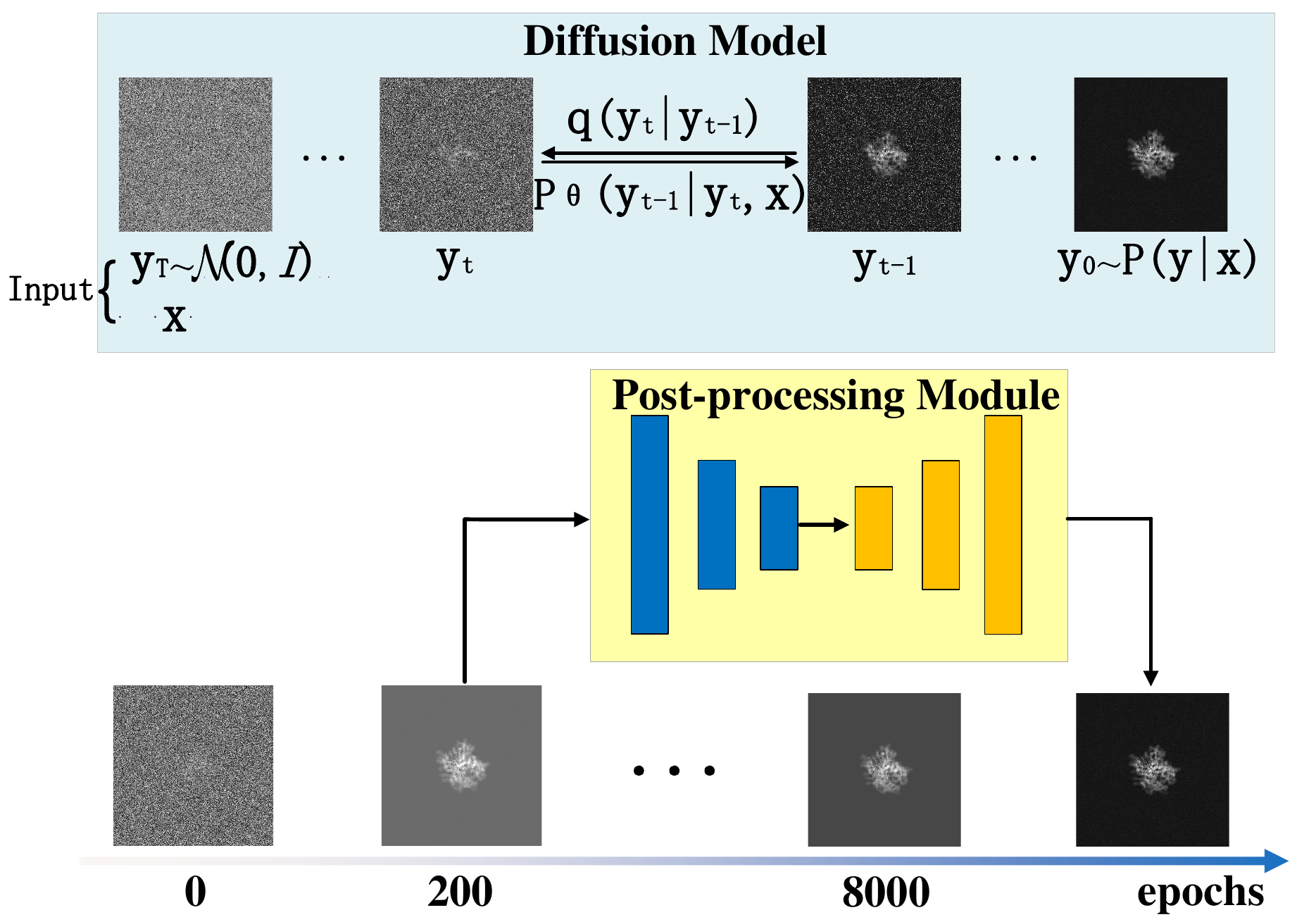} 
	\caption{Framework Overview. It includes two parts: diffusion model and post-processing module. The diffusion model restores the particle structure, and the post-processing accelerates the convergence.}
	\label{fig:2}
  \vspace{-0.4cm} 
\end{figure}

Single particle cryo-EM images denoising and restoration aims to remove all interference noise and restore the structure of the particle itself.~In this work, we not only need to excellently restore the real frozen state of particles under the interference of high background noise, but also collapse the indistinguishable structural noise covering the particle surface. Our proposed method, illustrated in the Fig.~\ref{fig:2}, involves a primary framework employing a diffusion model to restore the particle's structure, followed by a lightweight post-processing module to expedite model convergence. Below we provide a comprehensive description of our methodology.
\vspace{-0.05cm} 

\vspace{-0.2cm}
\subsubsection{Diffusion Model}
\label{sssec:subsubhead}
The initial diffusion model, like the Denoising Diffusion Probabilistic Models (DDPM), is an unconditional generative network that converts the  Gaussian distribution into a specific data distribution according to the process of iterative denoising. There are two processes: diffusion and reverse processes. The diffusion process is parameter-free, and the mean and variance are determined at each step. Given the initial data distribution $ y_{0}\sim q(y)$, the noise is continuously injected until it becomes an independent Gaussian distribution. In the reverse process, a model with parameters $p_{\theta}(y_{0})$ is used to learn to recover the initial data from Gaussian noise.

Due to the single particle restoration requires further denoising on cryo-EM images, unlike DDPM which starts from pure Gaussian noise. The noisy cryo-EM image $x$ can be used as prior conditions to guide model learning. Therefore, in this work we use a conditional diffusion model, Super-Resolution via Repeated Refinement (SR3) \cite{saharia2022image}. The goal is to learn a conditional model $p_{\theta}(y_{0}|x)$. So the reverse process can be expressed as:
\begin{equation}
	\begin{array}{c}
            p_{\theta}(y_{0:T}|x)= p_(y_{T})\prod_{t=1}^{T}p_{\theta}(y_{t-1}|y_{t},x)  
	\end{array}\
\end{equation}
\begin{equation}
	\begin{array}{c}
             p_{\theta}(y_{t-1}|y_{t},x)= \mathcal{N}(y_{t-1}|\mu_{\theta }(x,y_{t},\gamma _{t}) , \sigma _{t}^{2}I ) 
	\end{array}\
\end{equation}
where$(x,y_{0})$are sampled jointly from the data distribution. $\gamma$ is the variance of the added noise. Through conditioning on $\gamma$, the denoising model can understand the denoising level, which is conducive to the selection of the number of diffusion steps and the noise scheduling in the inference process.

\vspace{-0.2cm}
\subsubsection{Post-processing Module}
\label{sssec:subsubhead}
   
Diffusion model has shown outstanding performances, but it has a major drawback: it's gradual sampling characteristics will greatly extend the experimental time. In the experiment we observed that for single particle cryo-EM images denoising and restoration, the diffusion model can quickly reveal the particle structure, but it takes longer time to gradually reduce noise. However, the main problem that plagues current methods is the difficulty in identifying particle structures under high-noise conditions. Therefore, we first use the diffusion model to quickly figure out the particle structure (only trained for 200 epochs), and then we can use a current method to carry out step noise reduction on this basis. In our work, we use a simple U-Net \cite{ronneberger2015u} to complete this module. Through experimental observations, under the same hardware configuration, our framework can achieve better results and shorten the training time by nearly 12 times. The training time of our framework is less than 1 day, and the test results are shown in the table~\ref{tab:denoise}.~When the diffusion model iterates to 5,000 epochs, take 12 days, and the test results of EMD-8511 and EMD-24928 are respectively: MSE (1.028-2), PSNR (21.264), SSIM (0.824) and MSE(1.298e-2)  PSNR(20.163)   SSIM(0.870).

\vspace{-0.2cm}
\subsection{Training}
\label{ssec:subhead}
The training part of the method is divided into two steps. Firstly, for the diffusion model, the training goal is to optimize the log likelihood function $logp_{\theta}(y_{0}|x)$. According to SR3, the optimization objectives are simplified to:
\begin{equation}
	\begin{array}{c}
\mathbb{E}_{(x,y)} \mathbb{E}_{\epsilon ,\gamma } \left \| f_{\theta }(x,\sqrt{r} y_{0}+\sqrt{1-\gamma }\epsilon ,\gamma   ) -\epsilon  \right \|_{2}  
	\end{array}\
\end{equation}
where $\epsilon \sim \mathcal{N}(0,I)$, $f_{\theta }$ is the denoising model trained to predict noise vectors $\epsilon$ given any noisy image.

Secondly, for the training of the post-processing module, we use the mean squared error (MSE) loss function between the output images and the ground truth images.

\vspace{-0.3cm} 
\section{Experiment}
\label{sec:experiment}
\vspace{-0.2cm}
\subsection{Datasets}
\label{ssec:datasets}
In this section, we perform numerical experiments with (i) two simulated datasets and (ii) a real dataset.

 The simulated datasets is prepared from the particles of human HCN1 hyperpolarization-activated cyclic nucleotide-gated ion channel and dolphin prestin, which can be obtained in the Electron Microscopy Data Bank (EMDB) with the entries of EMD-8511 \cite{lee2017structures} and EMD-24928 \cite{bavi2021conformational} respectively. We first classify and remove noise in the 3D structure through the DeepTracer \cite{guan2022deeptracer}, and then generate clean 2D projection images in the direction of random sampling of the sphere. The simulated data is generated by adding Gaussian white noise to the clean projected image.

 The real dataset is a real-world experimental dataset of the human HCN1 hyperpolarization-activated cyclic nucleotide-gated ion channel. It can be available in the Electron Microscope Pilot Image Archive (EMPIAR) with 10081 entry.  


\vspace{-0.3cm}
\subsection{Experimental Setup}
\subsubsection{Experimental Scheme}
In this paper, all validation experiments are performed on the simulation datasets, and generalization experiments are conducted on the real dataset. Firstly, we compare the effect of denoising complex noise in different method. Then, we construct the 3D particle from the denoised images and compare them to the groundtruth. Finally, the well-trained model from the denoising model is generalized to the real EMPAIR particle dataset to verify the extensibility of the proposed method.

\vspace{-0.3cm}
\subsubsection{Evaluation Metrics}
As representative metrics in the field of image denoising and restoration, MSE, PSNR and SSIM are used to measure the results of denoising methods.~And then we conducted three dimensional reconstruction experiments on the denoising result.~The Fourier shell correlation~(FSC)~\cite{van2005fourier} measures the normalized crosscorrelation between two 3D maps in Fourier domain at various radial frequency shells. The resolution of the 3D reconstruction results can be defined as the spatial frequency at which the FSC curve is 0.143.

\vspace{-0.3cm}
\subsection{Experimental results}
We make a qualitative and quantitative comparison and analysis of our method with the current mainstream methods, including: (1) BM3D \cite{dabov2007image}, (2) K-SVD \cite{aharon2006k}, (3) Covariance Wiener Filter (CWF) \cite{bhamre2016denoising}, (4) DenoiseGAN \cite{su2018generative}, (5) RobustGAN \cite{gu2020generative}, (6) Noise2noise \cite{lehtinen2018noise2noise}. 

\begin{table}[tbp]
	\centering
	\caption{Denoising results in two simulated datasets: the Mean Square Error (MSE), the Peak Signal-to-Noise Ratio (PSNR) and the Structural Similaity Index Measure (SSIM) of different models. Our method achieves the best results.}
	\label{tab:denoise}
	\resizebox{\columnwidth}{!}{
		\begin{tabular}{c|c c c |c c c}
			\hline
			\multirow{2}{*}{Methods} & \multicolumn{3}{c|}{EMD-8511}  & \multicolumn{3}{c}{EMD-24928} \\ \cline{2-7}
			& MSE $\downarrow $ & PSNR $\uparrow$  & SSIM $\uparrow$ & MSE $\downarrow $ & PSNR $\uparrow$  & SSIM $\uparrow$ \\ \hline
			BM3D \cite{dabov2007image} & 1.323e-1 & 4.963 & 0.425 & 7.705e-2 & 7.419 & 0.635  \\ \hline
			K-SVD \cite{aharon2006k} & 1.351e-1 & 4.812 & 0.271 & 7.913e-2 & 7.240 & 0.404  \\ \hline
			CWF \cite{bhamre2016denoising}  & 1.319e-2 & 19.071 & 0.808 & 2.948e-2 & 15.354 & 0.652  \\ \hline	
			DenoiseGAN \cite{su2018generative} & 3.299e-3 & 20.693 & 0.885 & 4.088e-3 & 19.405 & 0.885  \\ \hline
			RobustGAN \cite{gu2020generative}& 3.239e-3 & 21.884 & 0.889 & 3.484e-3 & 19.569 & 0.902 \\ \hline
            Noise2noise \cite{lehtinen2018noise2noise}& 1.776e-3 & 24.846 & 0.913 & 2.513e-3 & 22.221 & 0.909  \\ \hline
			Ours & \textbf{1.449e-3} & \textbf{26.775} & \textbf{0.932} & \textbf{2.141e-3} & \textbf{23.356} & \textbf{0.923} \\ \hline
	\end{tabular}}
\end{table}

\begin{figure*}[h]
	\centering
	\includegraphics[width=\linewidth]{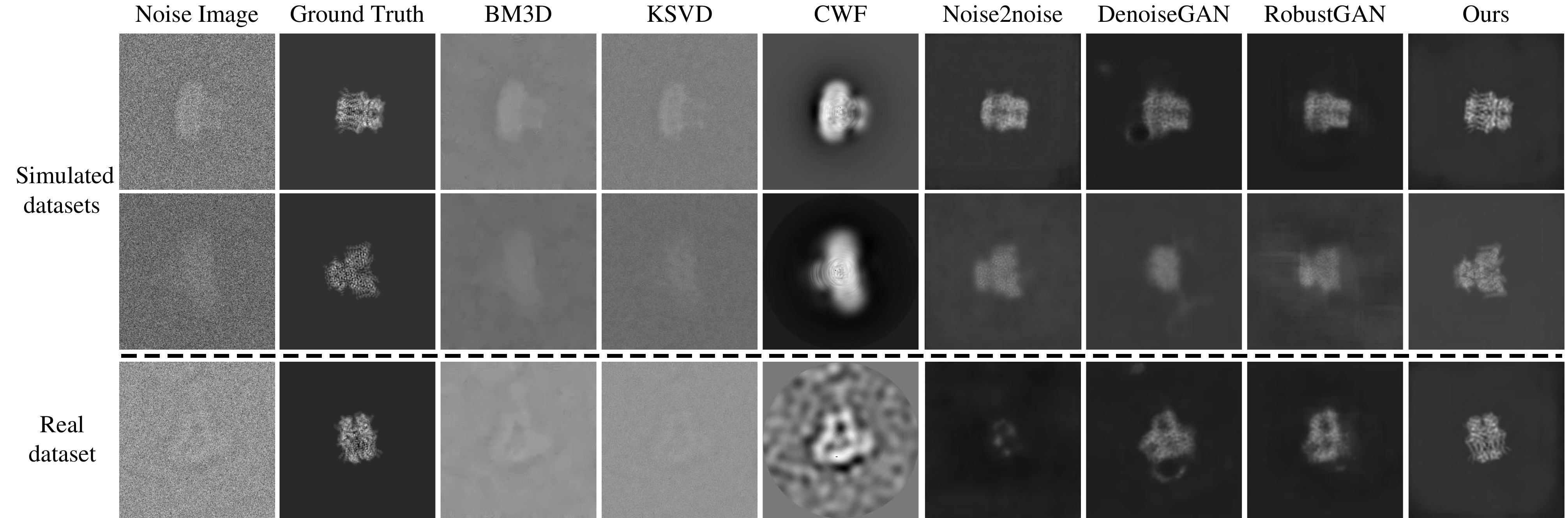} 
	\caption{Comparison of the denoising results from different methods. The first and second rows compare the denoising results of the EMD-8511 and EMD-24928 simulated datasets respectively, and the third row compares the denoising results in the real dataset of the EMPAIR-10081, whose ground truth is the closest projection map. Our method can effectively remove  unresolved structural noise on both simulated and real datasets under high noise conditions.}
	\label{fig:denoised}
  \vspace{-0.4cm} 
\end{figure*}

\begin{figure}[ht!]
	\centering
	\includegraphics[width=0.9\columnwidth]{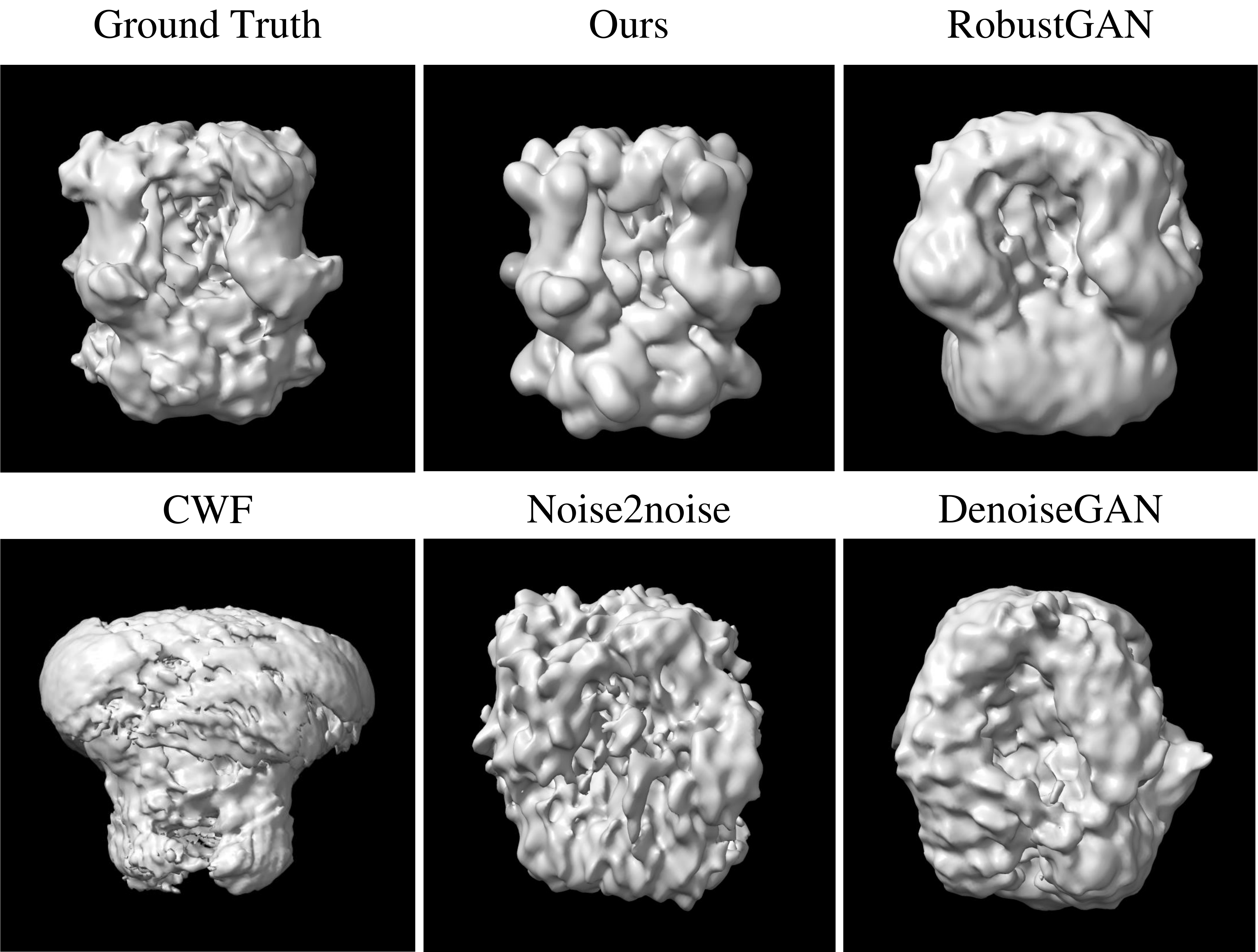} 
	\caption{The 3D reconstruction results of denoised images. Our results are closest to the ground truth.}
	\label{fig:3d}
 \vspace{-0.4cm} 
\end{figure}

Table~\ref{tab:denoise} presents a quantitative comparison of the denoised results. Our method outperforms others in terms of all three assessment criteria, demonstrating superior performance. In the first and second rows of Fig.~\ref{fig:denoised}, the samples of denoised images by different methods are displayed. We recognize traditional methods such as BM3D, KSVD, and CWF can extract the general outline of the single particle under high noise conditions, especially in CWF results, but they still leave the part of structural noise that is difficult to distinguish. The deep learning methods can perform much better, but they produce ambiguous boundaries. For the sake of fairness here, retraining is done on the datasets whose groundtruth is free of structural noise, which is not operated in their original algorithm. 

To verify the generalization and scalability of the proposed  method,  we conduct comparative experiments with real particle dataset. The specific experimental protocol is to use the optimal network model trained on the simulation dataset to test on the real dataset. As shown in Fig.~\ref{fig:denoised}, our method still shows the best denoising effect, while deep learning methods that perform well on simulated data, such as Noise2noise, DenoiseGAN, and RobustGAN, perform very poorly on real data. This also proves the practical value of our method.
\begin{figure}[ht!]
	\centering
	\includegraphics[width=0.9\columnwidth]{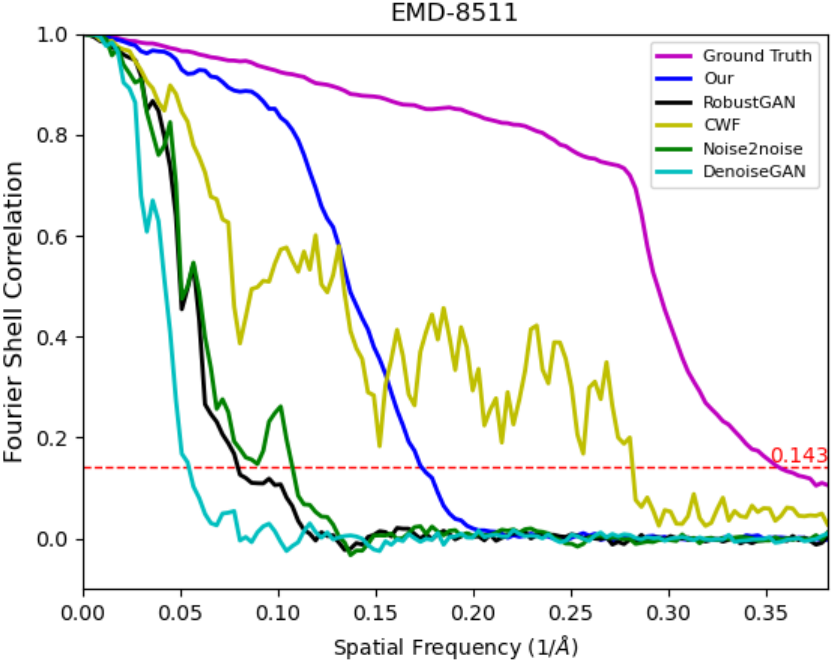} 
	\caption{FSC of the 3D reconstruction results. According to the 0.143 cutoff criterion (Dashed line), as the spatial frequency increases, the resolution of 3D structures gradually increases and the FSC value of highly reliable structures steadily approaches 0. Our results exhibits highly reliability and high-resolution.}
	\label{fig:fsc}
\vspace{-0.4cm} 
\end{figure}

In order to further prove the importance of denoising and verify the effect of restoration, we performed 3D reconstruction based on the denoised image using the RELION tool \cite{scheres2012relion}. 
The Fig.~\ref{fig:3d} shows the 3D reconstruction results of denoised images based on all algorithms. It can be seen that only our method restores a complete and clear structure (BM3D and K-SVD reconstruction failed).  

After reconstruction, we used FSC to determine the resolution and reliability of the reconstructed particle structure. As shown in the Fig.~\ref{fig:fsc} the resolution is determined by the 0.143 cutoff of FSC (Red dashed line). From left to right, the resolution of the particle structure gradually increases. Although our reconstructed result is not as good as the ground truth, it is better than other methods. In addition, a structure with high reliability should decay monotonically toward 0 as the spatial frequency increases (Abscissa). Obviously, the result of CWF is unreliable. Moreover, it can be seen from the Fig.~\ref{fig:3d} that the reconstructed result still has the structure caused by structural noise.


%
%
%


\vspace{-0.5cm} 
\section{CONCLUSIONS}
\label{sec:conclutions}
\vspace{-0.2cm}
In this study, we present a framework based on the diffusion model with post-processing to address the task of single-particle restoration in cryo-EM images. Extensive experimental results substantiate the feasibility and effectiveness of our proposed approach. Our method not only achieves the best denoising results on both simulated and real datasets, but also effectively reduces unresolved structural noise. 

\bibliographystyle{IEEEbib}
\bibliography{refs}

\end{document}